\documentclass[letterpaper, 10 pt, conference]{ieeeconf}  % Comment this line out if you need a4paper

\IEEEoverridecommandlockouts                              % This command is only needed if 
\usepackage{placeins} 
\usepackage{float}
\usepackage{cite}
\usepackage{amsmath,amssymb,amsfonts}
\usepackage{algorithmic}
\usepackage{graphicx}
\usepackage{textcomp}
\usepackage{xcolor}
\usepackage{utfsym}
\usepackage{multirow}
\usepackage{csquotes}

\usepackage{subcaption}
\usepackage{caption}
\usepackage{booktabs}
\usepackage{multirow}
\usepackage{xcolor}
\newcommand{\pub}[1]{\textcolor{gray}{\footnotesize #1}}

\title{\LARGE \bf
Traj-VLN: Learning Pixel-Space Interaction via Autoregressive Trajectory Generation
}

\author{Changfei Fu, Guangcheng Chen, Aoxiang Gu, Haoxiang Liang,\\
Wenjun Xu$\dagger$, and Hong Zhang$\dagger$, \textit{Life Fellow, IEEE} % <-this % stops a space
\thanks{$\dagger$Corresponding author (hzhang@sustech.edu.cn)}%
\thanks{
Changfei Fu and Hong Zhang are with the Shenzhen Key
Laboratory of Robotics and Computer Vision, Southern University of
Science and Technology, Shenzhen, China. Changfei Fu and Wenjun Xu are also with the Peng Cheng National Laboratory, Shenzhen, China. Weinan Chen is with the State Key Laboratory of Precision Electronic Manufacturing Technology and Equipment, Guangdong University of Technology, Guangzhou, China. This work was supported by the Shenzhen Key Laboratory of Robotics and Computer Vision (ZDSYS20220330160557001), the Major Key Project of PCL (PCL2024A04), and the National Natural Science Foundation of China under Grant
U21A20476.}%
}

\begin{document}

\maketitle
\thispagestyle{empty}
\pagestyle{empty}

%%%%%%%%%%%%%%%%%%%%%%%%%%%%%%%%%%%%%%%%%%%%%%%%%%%%%%%%%%%%%%%%%%%%%%%%%%%%%%%%
\begin{abstract}
Leveraging vision-language models (VLMs) to achieve generalization has become a prevailing paradigm in Vision-and-Language Navigation in Continuous Environments (VLN-CE). A VLN task can be decomposed into a sequence of 3D spatial interactions with the environment, as described by instructions such as \enquote{walk to the end of the sofa and turn left.} However, such spatial interactions, which involve moving into the image along the direction of depth sensing are puzzling for VLMs, since they are predominantly trained on conversations including only RGB images. Prior work has tackled this issue by fine-tuning VLMs with a pixel-goal as the supervision signal. But how can VLMs
accurately infer a target pixel if they lack
explicit understanding of the 3D scene layout? 

Rather than incorporating 3D geometric information (which VLMs rarely encounter during pre-training) or simply supervising the VLM using a single pixel-goal, we propose an alternative approach: fine-tuning VLMs to learn navigation interactions directly in 2D pixel space via autoregressive trajectory generation. Given a linguistic instruction and historical observations, our model sequentially predicts a series of pixel coordinates, drawing a trajectory from the bottom center of the current observation-a unique projection of the 3D trajectory. This autoregressive generation of pixel trajectory effectively serves as a chain-of-thought mechanism for more accurate pixel-goal prediction and enables the VLM to plan executable interactions in cluttered environments without explicit 3D reasoning. While prior work has proved that pixel-goal supervision outperforms learning of discrete actions, our experiments further verify that the pixel-trajectory supervision significantly enhances VLN-CE performance. Moreover, we demonstrate that our flagship model achieves substantial improvement over state-of-the-art methods that use similar scale of training data.
\end{abstract}

%%%%%%%%%%%%%%%%%%%%%%%%%%%%%%%%%%%%%%%%%%%%%%%%%%%%%%%%%%%%%%%%%%%%%%%%%%%%%%%%
\section{INTRODUCTION}
Vision-and-language navigation (VLN) requires an embodied agent to navigate in unseen environments following natural language instructions, while receiving continuous visual observations from onboard sensors. The instructions are typically composed of detailed sub-task descriptions-for example, “walk out of the dining room, into the living room, and take the first right into the recreation room; stop between the door and the pool table.” Unlike traditional navigation paradigms, which focus on moving the agent from one position coordinate to another, VLN emphasizes the process of 3D spatial interactions with the environment\cite{b1}. Driven by the powerful generalization capabilities of LLMs, employing VLMs to project visual information into language space has become a prevailing paradigm in VLN methods\cite{b2,b3,b4}.

\begin{figure}
  \centering
  \includegraphics[width=1\linewidth]{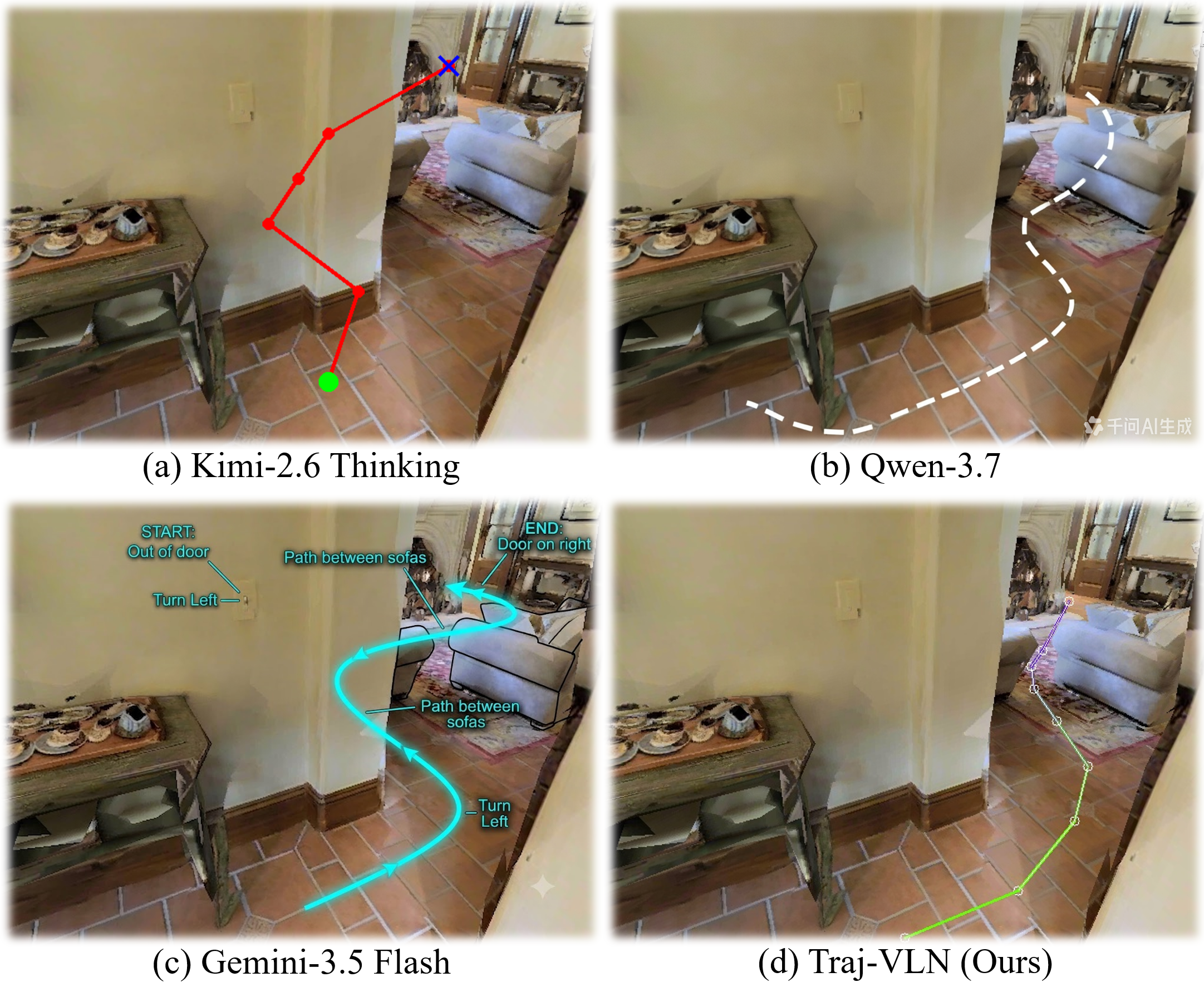}
  \vspace{-6mm}
  \captionsetup{font={small},labelfont=bf}
  \caption{Qualitative comparison of Traj-VLN with general-purpose multi-model large language models (MLLMs). It's demonstrated that the popular MLLMs can hardly draw a reasonable trajectory to interact with the 3D environments. With our fine-tuning supervised by pixel-space trajectories, the VLMs achieve the interactions in 2D image plane by learning to autoregressively generate a sequence of pixel coordinates. The prompt used for the general-purpose MLLMs is \enquote{please draw a trajectory on the image to follow the instruction: get out of the door and turn left, go through the way between the sofas, pass the fireplace to reach the door on the right.}}
  \label{fig1}
\vspace{-8mm}
\end{figure}

Existing VLN models mostly fine-tune VLMs supervised by direct action chunks\cite{b1,b2,b4,b5,b6}, which in fact compose a movement trajectory in the 3D space. However, existing VLMs and their vision encoders almost exclusively inherit the CLIP paradigm pre-trained on conversations with 2D images. Planning a 3D trajectory is inherently difficult without perceiving the 3D physical layout (Fig.~\ref{fig1}). Recent advances have attempted to incorporate depth information \cite{b1} or geometric priors \cite{b7} into the VLN framework, yet these approaches have not completely resolved the problem, as VLMs lack sufficient exposure to 3D information utilization during pretraining\cite{b8,b9}. Consequently, additional learnable encoders are often incorporated to align depth embeddings with RGB representations\cite{b7,b9}. To formulate VLN as a visual-grounding task compatible with pre-trained VLMs, InternVLA-N1\cite{b11} first fine-tunes the VLM using pixel-goal supervision within the 2D pixel space that is inherently familiar to VLMs. Through single-variable experiments, Goal2Pixel\cite{b12} further demonstrates that pixel-goal supervision substantially improves VLN performance over action-chunk prediction. The distribution mismatch between the large-scale internet pretraining corpus of VLMs and the robotics fine-tuning data can lead to catastrophic forgetting\cite{b10}, which may explain the remarkable success of pixel-goal supervision\cite{b3,b11,b12}. 

To follow an instruction, a robot need to interact with the environment, such as passing a dining table from the right. A pixel-goal is the projection of the target waypoint, which is the result of such a 3D interaction. However, a critical question arises: how can VLMs accurately infer a target pixel on the image if they lack explicit understanding of the 3D scene layout? Assuming that the VLMs do not inherently carry the ability of effective trajectory reasoning, we are motivated to train the VLMs with a explicit reasoning process of obtaining the resulting pixel-goal. We fine-tune VLMs using 2D trajectories paired with detailed instructions to teach VLMs the interaction processes. This paired supervision enables the model to learn the unambiguous 2D projections of 3D trajectories within the familiar 2D pixel space, circumventing the need for explicit 3D reasoning. We argue that the objective of VLM fine-tuning should be to reorganize and leverage the priors embedded in pre-training data, alongside the emerging reasoning capabilities. To enable the model to perform such interactions via drawing a trajectory from near to far, we fine-tune VLMs to autoregressively generate a sequence of pixel coordinates using a simple cross-entropy loss. For each sequence of observations in the training data, we project future robot poses onto the current observation to obtain the ground-truth of subsequent trajectory for supervision. Occluded and duplicate waypoints are filtered out using the depth image. The textual format of these coordinates aligns naturally with the pretraining corpus, ensuring the model understands the generated representations. The autoregressive mechanism, where the previous predictions serve as input for subsequent generation, drives the VLM to capture the sequential nature of the interaction processes, effectively functioning as a chain-of-thought (CoT) \cite{b13} for pixel-goal prediction.

In this paper, we propose Traj-VLN, which enables VLMs to learn pixel-space interactions and thereby achieves superior target pixel prediction. The contributions of this paper are summarized as follows: 1) We propose Traj-VLN, which fine-tunes general-purpose VLMs using paired supervision from 2D trajectories and language instructions, enabling the model to autoregressively generate the trajectory waypoints in the pixel space and thereby learn the underlying interaction processes. 2) Our experiments demonstrate that the VLN model supervised by 2D trajectories significantly outperforms its counterpart supervised by pixel-goal under identical settings. 3) By adopting the final point of the generated trajectory as the pixel-goal for evaluation, we reveal that autoregressive trajectory generation effectively serves as a CoT mechanism for better pixel-goal prediction.

\section{RELATED WORK}
\vspace{-2mm}
Our Traj-VLN address the spatial reasoning bottleneck in VLM-based VLN-CE by introducing 2D trajectories as the supervision signals to fine-tune VLMs in the pixel space, which is inherently aligned with the distribution of pre-training data. 
\vspace{-1.5mm}

\subsection{VLN-CE by Fine-tuning the Pretrained VLMs}
VLN has recently witnessed remarkable progress in simulation and benchmarking from nodes selecting in the discrete-defined nav-graph\cite{b14,b15,b16,b17} to more realistic VLN in continuous environments (VLN-CE), where the robot is allowed to navigate to any unobstructed locations \cite{b18,b19}. Further, to address the limitations in sim-to-real and cross-scene generalization caused by artificially customized neural networks and scarce training data \cite{b20,b21}, recent approaches have introduced general-purpose VLMs into this research field. Leveraging the powerful priors from large-scale pre-training data and the emerging commonsense reasoning capability of LLMs, VLM fine-tuning \cite{b2,b6,b4,b1,b11,b5,b3} has become a prevailing paradigm in VLN-CE.

As the first endeavour of VLM fine-tuning for VLN-CE, NaVid\cite{b2} inherits the main architecture of LLaMA-VID\cite{b22}, using four visual tokens and one instruction-queried visual token to represent historical observations. However, information redundancy is inevitable across consecutive visual observations\cite{b6}. We adopt an uniform down-sampling strategy for historical observations, which has been widely validated as an effective practice \cite{b4,b12,b1,b11,b3}.

Most existing VLN models \cite{b2,b5,b6,b4,b1} formulate the task as a Partially Observable Markov Decision Process (POMDP), where the VLM directly predict the next action based on the current observation, and the agent executes the predicted action to obtain a new observation. Navid-4D\cite{b7} further demonstrated that an action chunk of capacity 4 achieve the best performance. Such an action chunk, composed of primitive actions such as "move forward 25cm" and "turn left/right 15 degrees", constructs a trajectory in 3D space. However, VLMs and their vision encoders are predominantly pre-trained on conversations involving only 2D images, as datasets contaning 3D geometry or RGB-D images remain scarce. To bridge this gap, NaVid-4D and JanusVLN incorporated RGB-D images and 3D priors in geometric foundation models into their VLN frameworks, yet achieved limited improvements through aligning the 3D information with the pre-trained 2D space\cite{b1,b7,b8,b9}. By introducing pixel-goal as supervision for VLM fine-tuning, InternVLA-N1\cite{b3} outperforms these models that incorporate 3D information,  demonstrating that effective 2D-space supervision can surpass explicit 3D alignment. As the 3D spatial perception bottleneck becomes the focus of recent studies in VLM-based VLN-CE, we address this issue by propose a novel supervision paradigm in pixel space that is familiar to pretrained VLMs.

\begin{figure*}
  \centering
  \includegraphics[scale=0.288]{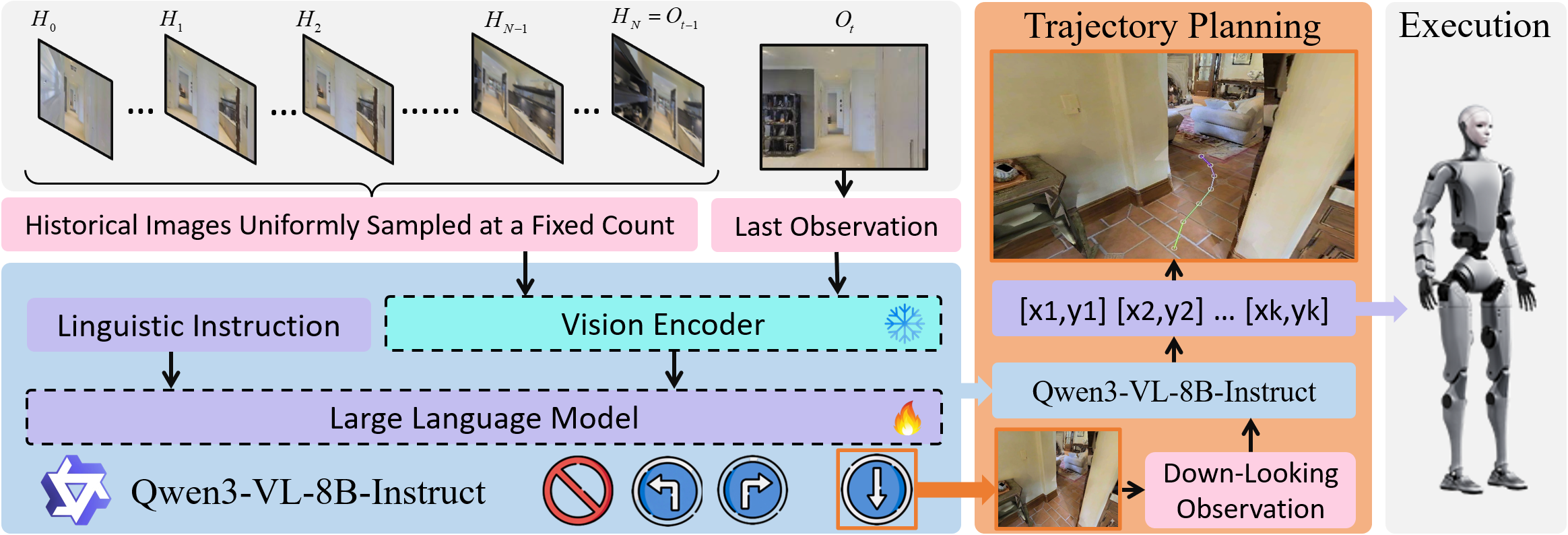}
  \vspace{-6mm}
  \captionsetup{font={small},labelfont=bf}
  \caption{The framework of Traj-VLN. At each time step, Traj-VLN takes as input a language instruction, the current observation, and a set of uniformly sampled historical observations. Based on Qwen3-VL, Traj-VLN fully fine-tunes the LLM and the projection layer while keeping the vision encoder frozen. In the first conversation stage , Traj-VLN is fine-tuned to output either turning actions or \enquote{STOP}. If Traj-VLN outputs \enquote{↓} in the first stage, it enters the second stage and receive an down-looking observation. During the second stage, Traj-VLN is fine-tuned to autoregressively generate a sequence of pixel coordinates on the down-looking image. Supervised by the pixel-space trajectories, Traj-VLN achieves interactions with the image regions by delineating a trajectory from near to far. }
  \label{Framework}
  \vspace{-7mm}
\end{figure*}

\subsection{Supervision Paradigms for VLMs in VLN-CE}

Currently, most VLN models supervise VLMs with low-level robot actions during fine-tuning\cite{b2,b5,b6,b4,b1}. While InternVLA-N1 \cite{b11} pioneered the practice of pixel-goal supervision by labeling the farthest waypoint in the pixel space, Goal2Pixel \cite{b12} further demonstrated through single-variable experiments that pixel-goal supervision outperforms low-level action chunk prediction. In our paradigm, a pixel trajectory is analogous to an action chunk, which consists of a sequence of discrete actions. An action chunk effectively constructs a 3D trajectory in the physical world, which account for semantic meaning in only the context of 3D layout. In contrast, each pixel in a pixel trajectory carries an inherent semantic meaning directly within the image. CoT drives VLMs to explicitly outputting the reasoning processes, which has been shown to significantly enhance both performance and interpretability. In our context, a pixel trajectory can serve as the reasoning process to obtain a target waypoint on the image.

To establish the trajectory supervision paradigm in the pixel space, our motivations are threefold: (1) A pixel trajectory serves as a unique projection of the 3D trajectory, inherently capturing the spatial interaction process described by the instruction. (2) Learning the 2D interactions in the pixel space inherently matches the pre-training data distribution. (3) Recognizing the spatial reasoning limitations of existing VLMs, we hypothesize that learning to autoregressively generate a trajectory can function as a CoT mechanism for pixel-goal prediction, enabling the model to decompose the final goal into sequential intermediate steps. 

\section{Method:Traj-VLN}
\subsection{Task Formulation}

Vision-and-Language Navigation in Continuous Environments (VLN-CE) requires an agent equipped with onboard sensors to navigate through unseen environments by following natural language instructions. These instructions typically consist of detailed descriptions of the navigation process, e.g., \enquote{Walk out of the dining room, into the living room, and take the first right into the recreation room. Stop between the door and the pool table.}

At time step $t$, in addition to the instruction, the agent receives the current RGB observation $O_t$ alongside the past observations $\boldsymbol{O_p}=\{O_1, O_2, \dots, O_{t-1}\}$.  To reduce computational complexity, $\boldsymbol{O_p}$ is often compressed via down-sampling\cite{b1,b11,b13} or token merging\cite{b2,b6}. Given these inputs, VLN models are expected to output language planning\cite{b24}, pixel-level goals\cite{b11,b12}, or even direct action chunks\cite{b7}. Consequently, a VLN model must comprehend the instruction, reason over historical observations to infer the remaining subtasks, and accurately predict the subsequent output conditioned on the current observation.

\subsection{Model Paradigm}
As shown in Fig. \ref{Framework}, we build Traj-VLN upon Qwen3-VL-8B-Instruct, a publicly available, general-purpose VLM. In all implementations, we keep the vision encoder frozen and fine-tune the LLM and the projection layer. We uniformly sample the historical observations $\boldsymbol{O}_{his}=\big\{H_1, H_2, ..., H_N\big\}\subseteq\boldsymbol{O}_p$ at a fixed count with dynamically adjusted intervals. Each image in $\boldsymbol{O}_{his}$ is resized to a smaller resolution (usually 392$\times$392), while the current observation $O_t$ is maintained at $640\times480$. Prior to model input, each image is further rounded so that its height and width are multiples of 28, complying with the merged patch size requirements of Qwen3-VL.

In the first stage of planning, we employ a chat template to constrain the VLM's output to either \enquote{STOP}, \enquote{$\downarrow$} or a sequence of the two directional symbols: \enquote{$\leftarrow$} or \enquote{$\rightarrow$}. If the agent has reached the target position when the VLM outputs \enquote{STOP}, the task is declared succeeded; otherwise failed. If the VLM outputs a sequence of \enquote{$\leftarrow$} or \enquote{$\rightarrow$}, the current conversation loop is restarted with updated vision input after executing this planning. If the VLM outputs \enquote{$\downarrow$}, the conversation get into the second stage with an additional down-looking observation incorporated as input. According to all the visual inputs, particularly the down-looking image at the current position, the VLM is trained to autoregressively generate a sequence of normalized pixel coordinates formatted as \enquote{$[u_1,v_1]\:[u_2,v_2]\:\dots\:[u_k,v_k]$}, to delineate a trajectory on the down-looking image starting from the bottom center. The conversation loop is also terminated and restarted after executing the trajectory planning. The alternating between planning and execution loops until the VLM outputs \enquote{STOP}. During executing the planned trajectory, the depth map corresponding to the down-looking image is required to transform these pixel coordinates into 3D physical waypoints.

\subsection{Navigation using Traj-VLN}

If the fine-tuned VLM outputs a sequence of direction symbols (e.g., \enquote{$\rightarrow\rightarrow\rightarrow$}) , the agent rotates by 15 degrees in the corresponding direction for each symbol. If the model output \enquote{STOP}, the agent terminates the task and stop there.

If the model outputs \enquote{$\downarrow$}, the conversation continues and it becomes the user's turn to reply. The agent lower its head for 30 degree to capture an down-looking view, which serves as the user's response. Subsequently, during the model's turn, it autoregressively generate a sequence of textual coordinates, which are parsed as pixel coordinates: $\boldsymbol{Traj}_{pixel}=\big\{(u_i, v_i)| i=1, 2, \dots, k\big\}$ to delineate a trajectory on the down-looking image. After executing the planned trajectory, the conversation with the VLM is restarted with updated visual observations.

Given the corresponding depth map of the down-looking view, the trajectory is extended to a sequence of 3D vectors:

\begin{equation}
\boldsymbol{Traj}_{depth}=\big\{(u_i,v_i,d_i)^\mathsf{T}|i=1,2,...,k\big\}
\end{equation}
Assuming the homogeneous camera intrinsics are represented by $\mathbf{K}$ and the camera extrinsics relative to the robot base by $\mathbf{T}_{c2b}$, each vector in $\boldsymbol{Traj}_{depth}$ is transformed into the robot frame (where $Z_i = d_i$) as follows:
\begin{equation}
\mathbf{K}=
\left( \begin{array}{cccc}
f_u & 0 & c_u & 0\\
0 & f_v & c_v & 0\\
0 & 0 & 1 & 0\\
0 & 0 & 0 & 1\\
\end{array} \right)
\end{equation}

\begin{equation}
(X_{i},Y_{i},Z_{i},1)^\mathsf{T}=d_{i}\mathbf{T}^{-1}_{c2b} \, \mathbf{K}^{-1} \, (u_{i},v_{i},1,\frac{1}{d_i})^\mathsf{T}
\end{equation}
\begin{equation}
\boldsymbol{Traj}_{3D}=\big\{(X_i,Y_i,Z_i)^\mathsf{T}|i=1,2,...,k\big\}
\end{equation}
With the 3D trajectory $\boldsymbol{Traj}_{3D}$ in the robot frame, the robot is controlled to iteratively move toward the closest 3D point until it is determined to have been reached. Once the current plan is deemed finished, the conversation is restarted.

\begin{figure}
  \centering
  \captionsetup{font={small},labelfont=bf}
  \includegraphics[scale=0.31]{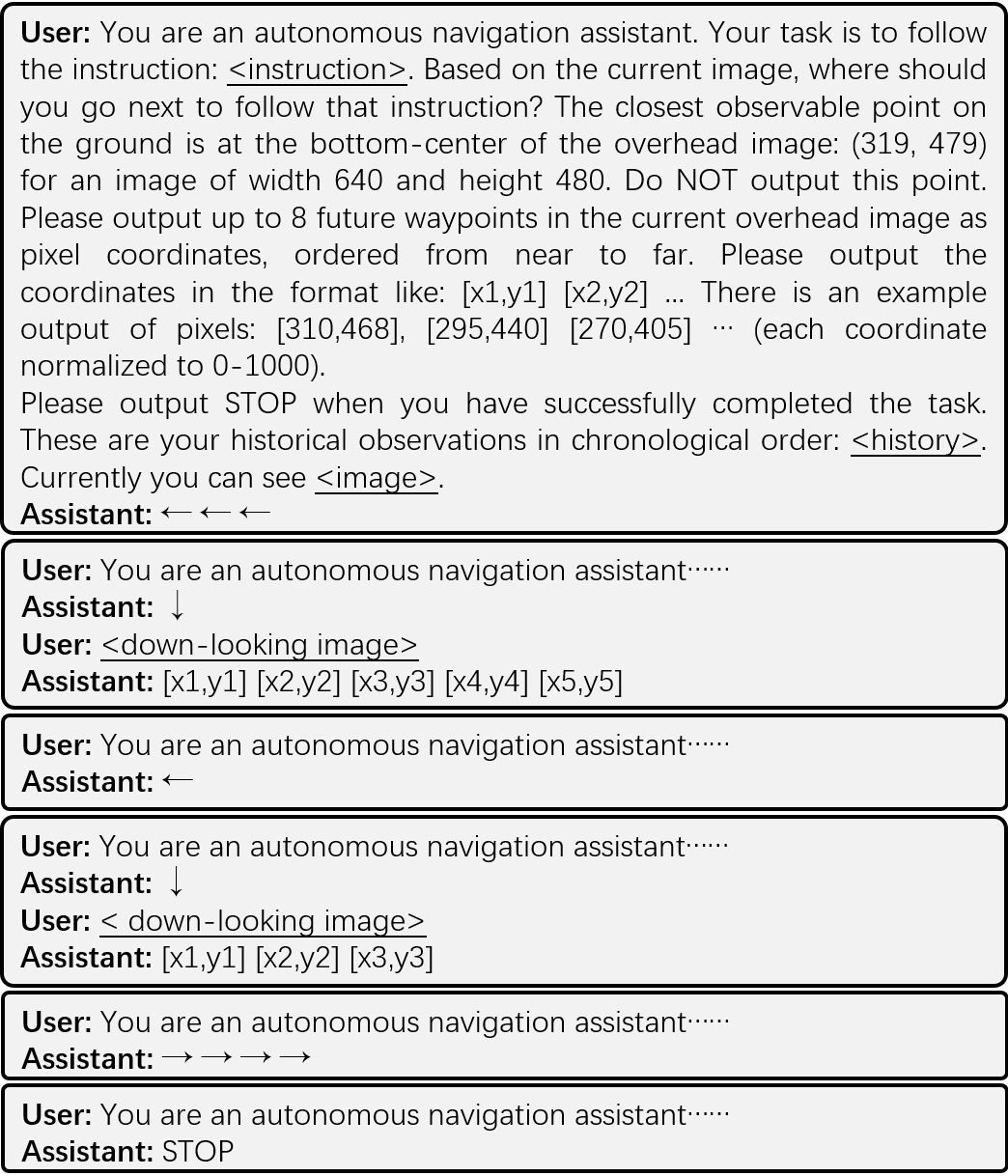}
  \vspace{-6mm}
  \caption{An example of conversations during a VLN task. The placeholders \enquote{ $<$instruction$>$}, \enquote{$<$image$>$}, \enquote{$<$down-looking image$>$}, and \enquote{$<$history$>$} are dynamically replaced with corresponding textual and visual content to form the input sequence, which is subsequently tokenized according to the VLM's vocabulary.}
  \vspace{-6mm}
  \label{chat_template}
\end{figure}

\subsection{Data Collection}

\begin{table*}[t]
\centering
\captionsetup{font={small}, labelfont=bf}
\caption{Comparison with SOTA methods on the VLN-CE R2R Val-Unseen (1,839 episodes) and RxR Val-Unseen
splits (3,669 episodes). The commonly used 2.6M training samples are derived from the standard R2R-CE (10.7K trajectories) and RxR-CE (19.5K trajecotries) datasets. InternVLA-N1 additionally employs a subset of the ScaleVLN dataset, comprising 77K trajectories. Following InternVLA-N1, we extract traning samples at an interval of 4 from these three datasets; however, we use a single robot height and view angles setting, yielding 1.75M samples, while InternVLA-N1 uses multiple robot height and angle configurations, resulting 4.72M samples. The training sample quantities for other methods are sourced from various additional datasets. To reduce training overhead, we freeze the vision encoder to save approximately 60\% of the training duration compared to full fine-tuning.}
\vspace{-2mm}
\label{tab:performance_comparison}
\resizebox{\textwidth}{!}{%  %
\begin{tabular}{l|l|l|l| cccc | cccc}
\toprule
\multirow{2}{*}{\shortstack[l]{Model\\Size}} & \multirow{2}{*}{\shortstack[l]{Vision\\Encoder}} & \multirow{2}{*}{\shortstack[l]{Training\\Samples}} & \multirow{2}{*}{Method} & \multicolumn{4}{c}{R2R-CE Val-Unseen} & \multicolumn{4}{c}{RxR Val-Unseen} \\
\cmidrule(lr){5-8} \cmidrule(lr){9-12}
 & & & & NE$\downarrow$ & OS$\uparrow$ & SR$\uparrow$ & SPL$\uparrow$ & NE$\downarrow$ & SR$\uparrow$ & SPL$\uparrow$ & nDTW$\uparrow$ \\
\midrule
\multicolumn{1}{c|}{$-$ } & \multicolumn{1}{c|}{$-$ } & \multicolumn{1}{c|}{$-$ } & Sim2Real \cite{b28} \pub{[CoRL24]} & 5.95 & 55.8 & 44.9 & 30.4 & 8.79 & 36.7 & 25.5 & 18.1 \\
\multicolumn{1}{c|}{$-$ } & \multicolumn{1}{c|}{$-$ } & \multicolumn{1}{c|}{$-$ } & AO-Planner \cite{b29} \pub{[AAAI25]} & 5.55 & 59.0 & 47.0 & 33.0 & 7.06 & 43.3 & 30.5 & 50.1 \\
\multicolumn{1}{c|}{$-$ } & \multicolumn{1}{c|}{$-$ } & \multicolumn{1}{c|}{$-$ } & NavMorph \cite{b30} \pub{[ICCV25]} & 5.75 & 56.9 & 47.9 & 33.2 & 8.85 & 30.8 & 22.8 & 44.2 \\
\midrule
7B & Unfrozen & 2.60M  & StreamVLN \cite{b6} \pub{[arXiv25]} & 5.43 & 62.5 & 52.8 & 47.2 & 6.72 & 48.6 & 42.5 & 60.2 \\
7B   & Frozen   & 3.55M & NaVid \cite{b2} \pub{[RSS24]} & 5.47 & 49.1 & 37.4 & 35.9 & 8.41 & 23.8 & 21.2 & $-$ \\
7B   & Frozen   & 1.84M & NaVid-4D \cite{b7} \pub{[ICRA25]} & 5.99 & 55.7 & 43.8 & 37.1 & $-$ & $-$ & $-$ & $-$ \\
7B        & Unfrozen & 15.6M & NaVILA \cite{b4} \pub{[RSS25]} & 5.22 & 62.5 & 54.0 & 49.0 & 6.77 & 49.3 & 44.0 & 58.8 \\
7B   & Frozen   & 2.60M  & JanusVLN \cite{b1} \pub{[ICLR26]} & 5.17 & 58.0 & 52.8 & 49.2 & 6.46 & 51.4 & 44.3 & 49.1 \\
7B   & Unfrozen & 4.72M & InternVLA-N1 \cite{b11} \pub{[ICLR26]} & 4.89 & 60.6 & 55.4 & 52.1 & 6.41 & 49.5 & 41.8 & 62.6 \\
8B   & Frozen   & 2.60M  & Goal2Pixel \cite{b12} \pub{[arXiv26]} & 4.80 & 58.3 & 53.9 & \textbf{52.7} & 6.91 & 48.1 & \textbf{44.7} & \textbf{63.0} \\
\midrule
8B     & Frozen   & 1.75M & Traj-VLN (ours) & \textbf{4.74} & \textbf{66.2} & \textbf{58.4} & 52.2 & \textbf{6.02} & \textbf{52.6} & 43.4 & 62.8 \\
\bottomrule
\end{tabular}% % <-- 注意这里的百分号
}
\vspace{-3mm}
\end{table*}

Following JanusVLN\cite{b1} and StreamVLN\cite{b6}, we construct our training data from the standard R2R-CE\cite{b25} (comprising approximately 10.7K trajectories and 152K samples) and RXR-CE\cite{b26} (comprising approximately 19.5K trajectories and 455.6K samples), supplemented with a subset of ScaleVLN\cite{b27} (comprising approximately 77.3K trajectories and 1.14M samples). Sampling at an action interval of 4 steps yields a total of 1.75M samples from approximately 107.5K trajectories. Each sample is represented as a tuple $(\boldsymbol{I},\boldsymbol{O}_{his},\boldsymbol{O}_{cur},\boldsymbol{Traj}_{pixel})$, where $\boldsymbol{I}$ denote the instruction, $\boldsymbol{O}_{his}$ is the historical observations, $\boldsymbol{O}_{cur}$ is the current observation, and $\boldsymbol{Traj}_{pixel}$ is the supervision signal for VLMs to predict.

To generate the supervision trajectory, we project the future poses relative to the current pose onto the current down-looking image. Let the poses of a $l$-length trajectory be denoted as:
\begin{equation}
\boldsymbol{Traj}=\big\{\mathbf{T}_{1},\mathbf{T}_{2},...,\mathbf{T}_{cur},...,\mathbf{T}_{l}\big\}
\end{equation}
where $\mathbf{T}_{cur}$ represents the pose of the current observation of a given sample:
\begin{equation}
\mathbf{T}_{cur}=\left( \begin{array}{cc}
\mathbf{R}_{cur} & \mathbf{t}_{cur} \\
0 & 1 \\
\end{array} \right)
\end{equation}
The future poses relative to $\mathbf{T}_{cur}$ are represented as:
\begin{equation}
\boldsymbol{Traj}_{future}=\big\{\mathbf{T}_i \mid cur<i \leq l \big\}
\end{equation}
We project the base positions of $\boldsymbol{Traj}_{future}$ onto the down-looking image as follows. First, we define the base pose:
\begin{equation}
\mathbf{T}^{base}_{i}=\left( \begin{array}{cc}
\mathbf{R}_{i} & \mathbf{t}_{i}-(0,0,height)^\mathsf{T} \\
0 & 1 \\
\end{array} \right)
\end{equation}
where \(\mathbf{t}^{base}_i = \mathbf{t}_i - (0,0,\text{height})^\mathsf{T}\) denotes the translation component. This base position is then projected onto the image plane:
\begin{equation}
(X_i,Y_i,Z_i,1)^{\mathsf{T}} =  \mathbf{K}\, \mathbf{T}^{-1}_{cur} \,(\mathbf{t}^{base}_{i}, 1)^\mathsf{T}
\end{equation}
yielding $n$ pixel coordinates:
\begin{equation}
\boldsymbol{Traj}_{pixel}=\big\{(u_i,v_i)=\frac{1}{{Z_i}}(X_i,Y_i)|i=1,2,...,n\big\}
\end{equation}
Finally we filter out invisible and occluded pixels from $\boldsymbol{Traj}_{pixel}$ and format the remaining into a textual string \enquote{$[u_1,v_1]\:[u_2,v_2]\:\dots\:[u_k,v_k]$}, which is incoporated into the chat template as the target for the VLM to predict.

\subsection{Training of Traj-VLN}

Our flagship Traj-VLN, based on Qwen3-VL-8B, was trained on a server with 8 Ascend 910B NPUs for approximately 91 hours, totally 728 NPU hours. Leveraging the 1.75M samples extracted from the three datasets, the training process comprised 16998 steps with a batch size of 128, and each step took approximately 23 seconds. The model was trained for one epoch, during which we exclusively fine-tuned the LLM and the projection layer with a learning rate of 2e-5, while keeping the vision encoder frozen. We observed that freezing the vision encoder saved approximately 62.5\% of the training time. Throughout all training and evaluation runs, the total number of historical images was fixed at 8. Except for the down-looking image, which was maintained at 640 × 480, the remaining frames were uniformly resized to 392 × 392. 

For the comparative experiments of Table \ref{tab:ablation_study} and Fig. \ref{final point as pixel-goal}, two models were trained under identical configurations using Qwen2.5-VL-7B as the backbone, with the LLM fine-tuned via LoRA (rank=32). To ensure a fair comparison between our method and those using pixel-goal supervision and prediction, we kept the same data, backbone and training settings across models, while varying only the supervision signals. On a server equipped with 4 Nvidia 4090 GPUs (each with 48GB of memory), each training run consumed approximately 171 hours to complete 274,266 steps.

\begin{table*}[t]
\centering
\captionsetup{font={small}, labelfont=bf}
\caption{Comparison of trajectory supervision and pixel-goal supervision on the VLN-CE R2R Val-Unseen and RxR Val-Unseen
splits. Two models are fine-tuned under identical training configurations, differing only in the supervision signal. \enquote{Pixel-Goal Evaluation} employs the shortest-path follower from the Habitat simulator to execute the predicted pixel goal. \enquote{Pixel-Goal Evaluation} for Traj-VLN treats the final pixel coordinate of the predicted trajectory as the pixel goal and evaluates it using the same pipeline as Pixel-Goal Evaluation. \enquote{Trajectory Evaluation} refer to sequentially executing the predicted waypoints in order.}
\vspace{-2mm}
\label{tab:ablation_study}
\resizebox{\textwidth}{!}{%
\begin{tabular}{l|l|l|l| cccc | cccc}
\toprule
\multirow{2}{*}{\shortstack[l]{Model\\Size}} & \multirow{2}{*}{\shortstack[l]{Vision\\Encoder}} & \multirow{2}{*}{\shortstack[l]{Training\\Samples}} & \multirow{2}{*}{\shortstack[l]{Method\\{\footnotesize (Fine-tuned by LoRA, Rank 32)}}} & \multicolumn{4}{c}{R2R-CE Val-Unseen} & \multicolumn{4}{c}{RxR Val-Unseen} \\
\cmidrule(lr){5-8} \cmidrule(lr){9-12}
 & & & & NE$\downarrow$ & OS$\uparrow$ & SR$\uparrow$ & SPL$\uparrow$ & NE$\downarrow$ & SR$\uparrow$ & SPL$\uparrow$ & nDTW$\uparrow$ \\
\midrule
7B & Frozen & 1.75M & \shortstack[l]{InternVLA-N1\\(Pixel-Goal Evaluation)} & 5.81 & 45.6 & 39.6 & 36.3 & 8.68 & 36.09 & 30.87 & 51.42 \\
\midrule
7B & Frozen & 1.75M & \shortstack[l]{Traj-VLN\\(Pixel-Goal Evaluation)} & 5.79 & \textbf{49.0} & 43.2 & 40.3 & \textbf{7.66} & \textbf{41.54} & \textbf{35.95} & \textbf{56.12} \\
\midrule
7B & Frozen & 1.75M & \shortstack[l]{Traj-VLN\\(Trajectory Evaluation)} & \textbf{5.66} & 48.3 & \textbf{43.6} & \textbf{40.4} & 7.98 & 39.38 & 33.76 & 48.71 \\
\bottomrule
\end{tabular}%
}
\vspace{-5mm}
\end{table*}

\section{Experiments}
\vspace{-1mm}
Our experiments are designed to investigate the following questions for evaluating Traj-VLN:
\begin{itemize}
\item (1) How well does Traj-VLN perform against SOTA VLN models under a comparable scale of training data?
\item(2) To what extent does pixel trajectory supervision outperform pixel-goal supervision in VLN?
\item(3) Can autoregressive trajectory generation serve as a chain-of-thought for improving pixe-goal prediction?
\item(4) Can Traj‑VLN learn to interact reasonably with the environment in pure pixel space?
\item(5) Does Traj‑VLN generalize effectively to unseen real‑world scenarios?

\end{itemize}

Our comparative experiments are conducted on the prominent VLN-CE benchmarks comprising 1839 R2R-CE val-unseen trajectories and 3669 RxR-CE val-unseen trajectories. We report the performance using standard VLN-CE metrics consisting of Navigation Error (NE), Oracle Success Rate (OS), Success Rate (SR) Success-weighted Path Length (SPL), and normalized Danymic Time Warping (nDTW).

\vspace{-1mm}
\subsection{Implementation and Experimental Setup}\label{AA}

To compare with SOTA methods on the VLN-CE benchmark, we build our flagship model on Qwen3-VL-8B. The model is trained for one epoch on a server with 8 Ascend 910B NPUs, taking approximately 91 hours. We fully fine-tune the LLM and the projection layer with a learning rate of 2e-5, while keeping the vision encoder frozen-a strategy that reduces approximately 62.5\% of the training duration. Our training data is collected from the standard R2R-CE, RxR-CE and the commonly-used ScaleVLN dataset. Sampling at intervals of 4 steps along the trajectories, we project the future poses onto the image plane and filter out invisible and duplicate coordinates. From the 107.6K trajectories across these three datasets, we construct a total of 1.75M training samples, each consisting of an instruction, historical images, the current image, and one of the following: a sequence of pixel coordinates, several turning actions, or \enquote{STOP}.

To compare the supervision of pixel-space trajectory with pixel-goal supervision, we trained two models under identical configurations while varying only the supervision signals, denoted as Traj-VLN and InternVLA-N1-S2 shown in Table \ref{tab:ablation_study}. Each model is based on Qwen2.5-VL and fine-tuned with LoRA at a rank of 32, with the vision encoder frozen. The models are trained for one epoch on a server with 4 Nvidia 4090 GPUs, taking approximately 171 hours. The training data for these two models are largely identical, differing only in the textual sequence for VLMs to predict: a single pixel coordinate indicating the target position versus a series of pixel coordinates to delineating a trajectory. The data used for these two models are the same 1.75M samples as those used for the flagship model.

\begin{figure}
  \centering
  \includegraphics[scale=0.3]{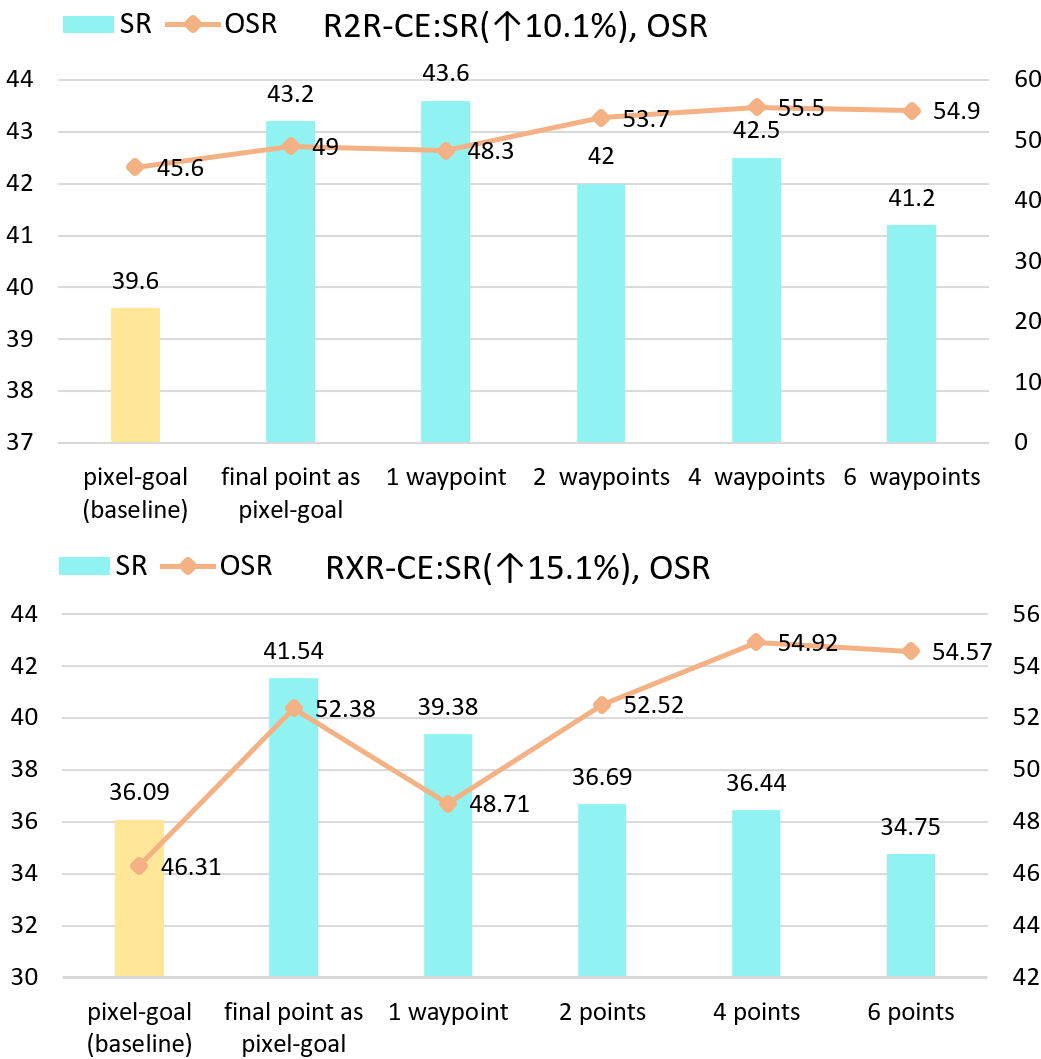}
  \vspace{-1mm}
  \captionsetup{font={small}}
\caption{Comparison of pixel-goal (InternVLA-N1) and Traj-VLN under varying numbers of execution steps per trajectory prediction. The horizontal axis denotes the number of waypoints along a trajectory to be executed. "Final pixel as pixel-goal" refers to treating the final waypoint of the predicted trajectory as the pixel goal for evaluating Traj-VLN using the same evaluation protocol as pixel-goal. The results are obtained with the same models as those reported in Table~\ref{tab:ablation_study}.}
  \vspace{-5mm}
  \label{final point as pixel-goal}
\end{figure}

\subsection{Results on VLN-CE Benchmark}\label{SCM}

As presented in Table \ref{tab:performance_comparison}, to address the research question (1), we compare our flagship Traj-VLN with seven VLM-based methods and three top-performing traditional approaches that do not employ VLMs. NaVILA, InternVLA-N1, Goal2Pixel, and our Traj-VLN all adopt 8 uniformly sampled images as the historical observations. Among the SOTA methods with a comparable scale of training data, Traj-VLN achieves the highest SR on both evaluation splits, despite using less training data.

JanusVLN and NaVid-4D are recent methods that incorporate 3D information to compensate for VLMs' inherent weakness in 3D perception. Traj-VLN surpasses JanusVLN by 5.6 and 1.2 points in SR on the two splits, respectively. Goal2Pixel and InternVLA-N1 are recent methods that adopt pixel-goal as the supervision signal. Despite using a more recent VLM backbone than InternVLA-N1, Goal2Pixel leverages less training data and keeps the vision encoder frozen. Unfreezing the vision encoder can lead to performance gains, but typically requires more training data and computational resources to mitigate overfitting. In our approach, we select Qwen3-VL as the backbone, which features an updated vision encoder, and keep it frozen during training. On the R2R-CE split, Traj-VLN achieves 58.4 in SR, improving by 3 points over InternVLA-N1. On the RxR split, Traj-VLN attains 52.6 in SR, outperforming InternVLA-N1 by 2.9 points. In summary, Traj-VLN consistently surpasses both methods that rely on different supervision signals and those that incorporate explicit 3D information to address the 3D perception challenge in VLM-based VLN-CE.

\begin{figure*}[h]
    \centering
    % 第一幅子图
    \begin{subfigure}{1\textwidth}
        \centering
        \includegraphics[width=\linewidth]{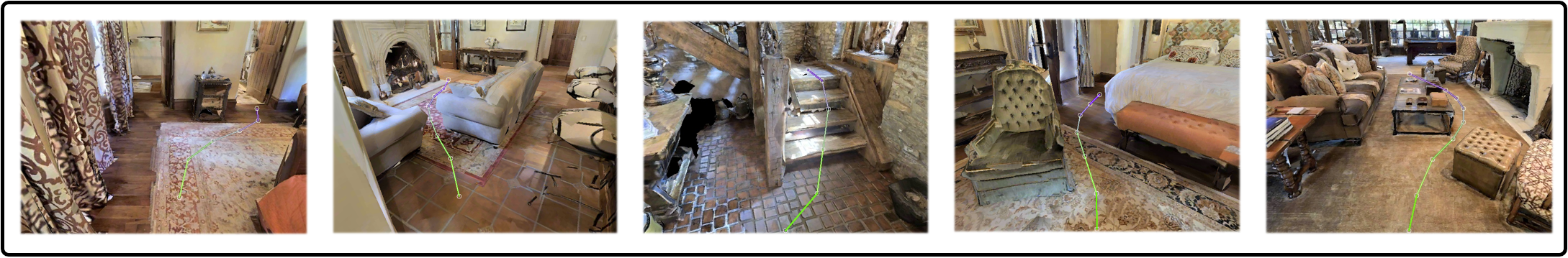}
        \vspace{-6mm}
        \caption{Planning results of Traj-VLN in the habitat simulation environments.}
        \label{qualitative-a}
    \end{subfigure}
    \vspace{0.3cm}  % 上下两幅图之间的垂直间距（可调）
    % 第二幅子图
    \begin{subfigure}{1\textwidth}
        \centering
        \includegraphics[width=\linewidth]{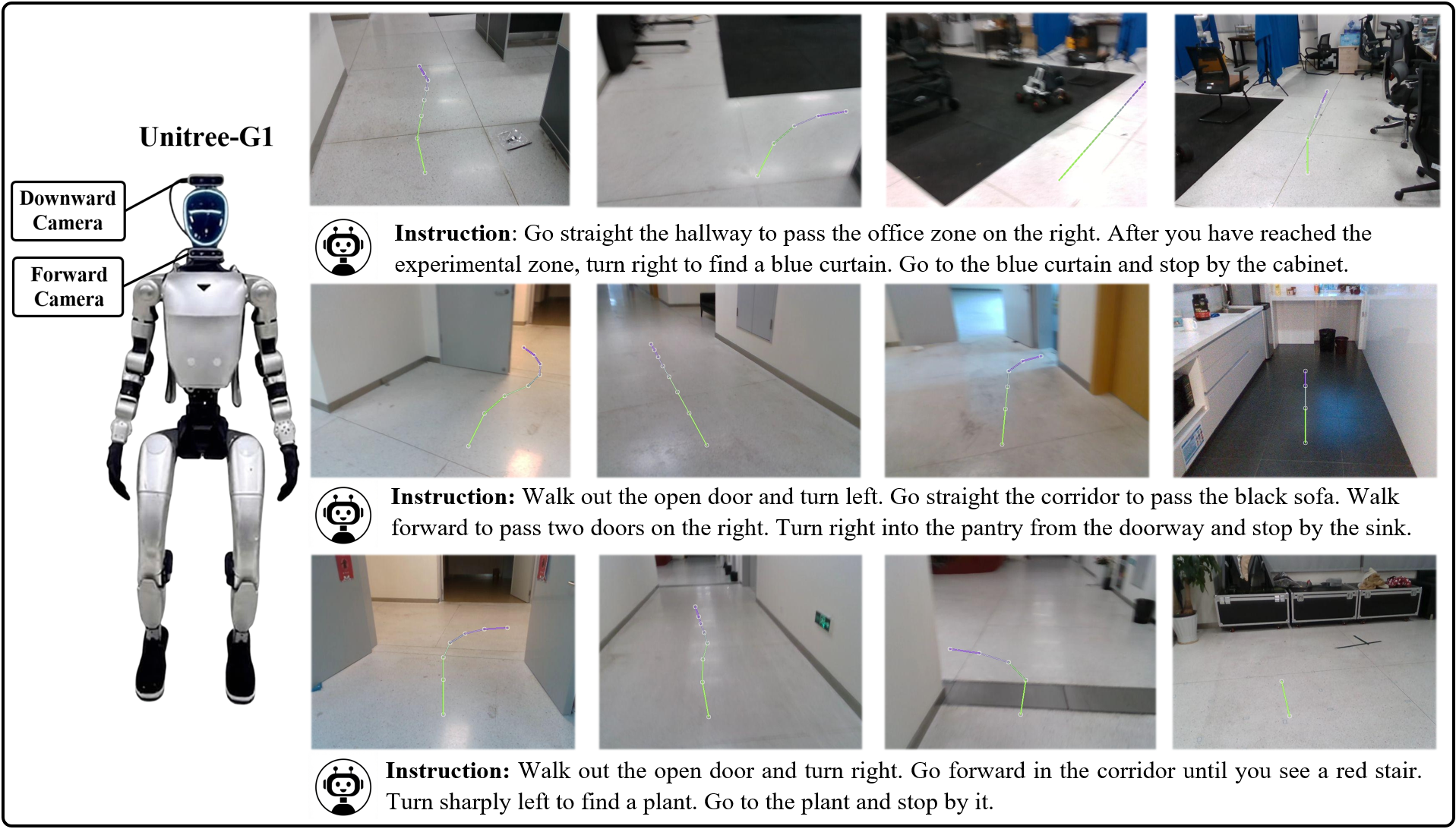}
        \vspace{-6mm}
        \caption{Planning results of Traj-VLN during three real-world VLN tasks.}
        \label{qualitative-b}
    \end{subfigure}
    \vspace{-9mm}
    \caption{Qualitative results of Traj-VLN in simulation and real-world environments. The Unitree-G1 robot is equipped with two cameras: a downward camera on the head for down-looking views and a forward camera positioned at the neck.}
    \vspace{-6mm}
    \label{qualitative}
    
\end{figure*}

\subsection{Comparison of Traj-VLN with Pixel-goal Supervision}

For question (2), we trained two models under identical configurations while varying only the supervision signals. Since Traj-VLN fine-tunes the VLM to predict a sequence of pixel coordinates, we encounter a trade-off problem during evaluation regarding how many waypoints to execute per trajectory prediction. To study the impact of this factor, we evaluate Traj-VLN with varying numbers of waypoint executed per prediction. As the overall trends shown in the Fig. \ref{final point as pixel-goal}, executing fewer waypoints per prediction-thereby increasing the reasoning frequency-leads to performance gains in SR, albeit at the cost of lower OSR. The chart also indicates that, regardless of the number of points executed, Traj-VLN consistently outperforms pixel-goal supervision in both SR and OSR. Among all configurations, executing only one waypoint per prediction (corresponding to a movement of approximately 0.5m) yields the highest SR. Under this setting, Traj-VLN surpass pixel-goal supervision by 4 and 3.29 in SR on the R2R-CE Val-Unseen and RxR-CE Val-Unseen splits, respectively. To compare more comprehensively, we report additional metrics in Table II, where the entries under \enquote{Final-Pixel Evaluation} correspond to executing one waypoint per prediction.

\subsection{Traj-CoT Validation}

Traj-VLN is built upon the hypothesis that learning spatial interaction processes over image regions yields greater VLN performance than supervising the VLMs solely with the resulting pixel-goal. Without such process-level supervision, the predicted pixel-goal may merely reflect a distribution learned from the training data, rather than a result of reasonable navigation actions. To verify the Traj-CoT mechanism, we adopt the final pixel coordinate along the trajectory predicted by Traj-VLN as the pixel goal for evaluation. The corresponding results are presented in Table \ref{tab:ablation_study} and Figure \ref{final point as pixel-goal}, where they are referred to as \enquote{Pixel-Goal Evaluation} and \enquote{final point as pixel-goal} respectively. In both \enquote{Pixel-Goal Evaluation} and \enquote{Final-Pixel Evaluation}, each conversation is terminated and restarted upon reaching the maximum of 8 movement steps. The results indicate that the final point of the trajectory generated by Traj-VLN serves as a better target than the model trained with pixel-goal supervision.

\subsection{Qualitative Results in Simulation and Real-World}
In this section we address questions (4) and (5) by presenting qualitative results in both simulation and the real world. As shown in Fig. \ref{qualitative-a}, we showcase several planning instances in cluttered scenes to demonstrate that Traj-VLN can interact reasonbly with the environment, even under challenging conditions such as stairs and clutterd spaces. Fig. \ref{qualitative-b} presents two real-world experiments conducted on a Unitree-G1 robot. Despite the captured images exhibiting blur and reflective artifacts, Traj-VLN consistently produces reasonable and collision-free trajectories. The results indicate that the planning capability of Traj-VLN generalizes well to real-world environments without relying on real-world training data.

\section{Conclusion}

We propose Traj-VLN, the first VLN framework that enables general-purpose VLMs to learn pixel-space interactions via autoregressive trajectory generation. We divide the conversation with the VLM into two stages: the VLM outputs turning actions in the first stage and shifts to the second stage if the output is "↓". In the second stage, the VLM is provided with an overhead image, on which it is supervised to delineate a trajectory by autoregressively outputting a series of pixel coordinates. 

By learning interaction processes directly in pixel space, Traj-VLN outperforms methods that supervise VLMs with a single pixel-goal coordinate and achieves state-of-the-art performance on the VLN-CE benchmark compared with the approaches using a similar scale of training samples. Through treating the final pixel coordinate of the predicted trajectory as the pixel-goal, we further demonstrate that our trajectory supervision effectively serves as a chain-of-thought mechanism for better pixel-goal prediction. Our qualitative results also show Traj-VLN's capability of correct interactions with cluttered environments, which is not readily available in pretrained VLMs. Moreover, real-robot experiments confirm that Traj-VLN generalizes well to unseen physical environments without real-world training data.


\vspace{-1mm}
\begin{thebibliography}{00}
\vspace{-2mm}

\bibitem{b1} Shuang Zeng, Dekang Qi, Xinyuan Chang, Feng Xiong, Shichao Xie, Ning Guo, et al., "Janusvln: Decoupling semantics and spatiality with dual implicit memory for vision-language navigation," in ICLR, 2026.


\bibitem{b2} Jiazhao Zhang, Kunyu Wang, Rongtao Xu, Gengze Zhou, Yicong Hong, He Wang, et al., "Navid: Video-based vlm plans the next step for vision-and-language navigation," in RSS, 2024.

\bibitem{b3} Meng Wei, Chenyang Wan, Jiaqi Peng, Xiqian Yu, Yuqiang Yang, Delin Feng, Wenzhe Cai et al. "Ground slow, move fast: A dual-system foundation model for generalizable vision-and-language navigation." ICLR, 2026.


\bibitem{b4} An-Chieh Cheng, Yandong Ji, Zhaojing Yang, Zaitian Gongye, Xueyan Zou, Jan Kautz, Erdem Bıyık, Hongxu Yin, Sifei Liu, and Xiaolong Wang. "Navila: Legged robot vision-language-action model for navigation." RSS, 2025.

\bibitem{b5} Jiazhao Zhang, Kunyu Wang, Shaoan Wang, Minghan Li, Haoran Liu, Songlin Wei, Zhongyuan Wang, Zhizheng Zhang, and He Wang. "Uni-navid: A video-based vision-language-action model for unifying embodied navigation tasks." arXiv, 2024.

\bibitem{b6} Meng Wei, Chenyang Wan, Xiqian Yu, Tai Wang, Yuqiang Yang, Xiaohan Mao, Chenming Zhu et al. "Streamvln: Streaming vision-and-language navigation via slowfast context modeling." arXiv, 2025.

\bibitem{b7} Haoran Liu, Weikang Wan, Xiqian Yu, Minghan Li, Jiazhao Zhang, Bo Zhao, Zhibo Chen, Zhongyuan Wang, Zhizheng Zhang, and He Wang. "NaVid-4D: Unleashing Spatial Intelligence in Egocentric RGB-D Videos for Vision-and-Language Navigation." ICRA, 2025.

\bibitem{b8} Peizheng Li, Zhenghao Zhang, David Holtz, Hang Yu, Yutong Yang, Yuzhi Lai, Rui Song, Andreas Geiger, and Andreas Zell. "Spacedrive: Infusing spatial awareness into vlm-based autonomous driving." CVPR, 2026.

\bibitem{b9}  Pingyi Chen, Yujing Lou, Shen Cao, Jinhui Guo, Lubin Fan, Yue Wu, Lin Yang, Lizhuang Ma, and Jieping Ye. "SD-VLM: Spatial Measuring and Understanding with Depth-Encoded Vision-Language Models." NeuIPS, 2026.

\bibitem{b10} Asher J. Hancock, Xindi Wu, Lihan Zha, Olga Russakovsky, and Anirudha Majumdar. "Actions as language: Fine-tuning vlms into vlas without catastrophic forgetting." ICLR, 2025.

\bibitem{b11} InternNav Team. "InternVLA-N1: An Open Dual-System Navigation Foundation Model with Learned Latent Plans." arXiv, 2025.

\bibitem{b12} Muyi Bao, Yuxin Cai, Hang Xu, Zongtai Li, Jinxi He, Jingfan Tang, Chen Lv, Ji Zhang, Yaqi Xie, and Wenshan Wang. "Goal2Pixel: Grounding Goals to Pixels for Vision-Language Navigation." arXiv, 2026.

\bibitem{b13} Shuo Wang, Yongcai Wang, Wanting Li, Xudong Cai, Yucheng Wang, Maiyue Chen, Zhizhong Su, Deying Li, and Zhaoxin Fan. "Aux-think: Exploring reasoning strategies for data-efficient vision-language navigation." NeurIPS, 2026.

\bibitem{b14} Duo Zheng, Shijia Huang, Lin Zhao, Yiwu Zhong, and Liwei Wang. "Towards learning a generalist model for embodied navigation." CVPR, 2024.

\bibitem{b15} Angel Chang, Angela Dai, Thomas Funkhouser, Maciej Halber, Matthias Niessner, Manolis Savva, Shuran Song, Andy Zeng, and Yinda Zhang. "Matterport3d: Learning from rgb-d data in indoor environments." 3DV, 2017.

\bibitem{b16} Peter Anderson, Qi Wu, Damien Teney, Jake Bruce, Mark Johnson, Niko Sünderhauf, Ian Reid, Stephen Gould, and Anton Van Den Hengel. "Vision-and-language navigation: Interpreting visually-grounded navigation instructions in real environments." CVPR, 2018.

\bibitem{b17} Alexander Ku, Peter Anderson, Roma Patel, Eugene Ie, and Jason Baldridge. "Room-across-room: Multilingual vision-and-language navigation with dense spatiotemporal grounding." EMNLP, 2020.

\bibitem{b18} Jacob Krantz, Erik Wijmans, Arjun Majumdar, Dhruv Batra, and Stefan Lee. "Beyond the nav-graph: Vision-and-language navigation in continuous environments." ECCV, 2020.

\bibitem{b19} Manolis Savva, Abhishek Kadian, Oleksandr Maksymets, Yili Zhao, Erik Wijmans, Bhavana Jain, Julian Straub et al. "Habitat: A platform for embodied ai research." ICCV, 2019.

\bibitem{b20} Yicong Hong, Zun Wang, Qi Wu, and Stephen Gould. "Bridging the gap between learning in discrete and continuous environments for vision-and-language navigation." CVPR 2022.

\bibitem{b21} Jacob krantz, Aaron Gokaslan, Dhruv Batra, Stefan Lee, and Oleksandr Maksymets. "Waypoint models for instruction-guided navigation in continuous environments." ICCV, 2021.


\bibitem{b22} Yanwei Li, Chengyao Wang, and Jiaya Jia. "Llama-vid: An image is worth 2 tokens in large language models." ECCV, 2024.

\bibitem{b23} Shaoan Wang, Yuanfei Luo, Xingyu Chen, Aocheng Luo, Dongyue Li, Chang Liu, Sheng Chen, Yangang Zhang, and Junzhi Yu. "VLingNav: Embodied Navigation with Adaptive Reasoning and Visual-Assisted Linguistic Memory." arXiv, 2026.

\bibitem{b24} Duo Zheng, Shijia Huang, Lin Zhao, Yiwu Zhong, and Liwei Wang. "Towards learning a generalist model for embodied navigation." CVPR, 2024.


\bibitem{b25} Jacob Krantz, Erik Wijmans, Arjun Majumdar, Dhruv Batra, and Stefan Lee. "Beyond the nav-graph: Vision-and-language navigation in continuous environments." ECCV, 2020.

\bibitem{b26} Alexander Ku, Peter Anderson, Roma Patel, Eugene Ie, and Jason Baldridge. "Room-across-room: Multilingual vision-and-language navigation with dense spatiotemporal grounding." EMNLP, 2020.

\bibitem{b27} Zun Wang, Jialu Li, Yicong Hong, Yi Wang, Qi Wu, Mohit Bansal, Stephen Gould, Hao Tan, and
Yu Qiao. Scaling data generation in vision-and-language navigation. ICCV, 2023.

\bibitem{b28} Zihan Wang, Xiangyang Li, Jiahao Yang, Yeqi Liu, and Shuqiang Jiang. "Sim-to-real transfer via 3d feature fields for vision-and-language navigation." arXiv, 2024.

\bibitem{b29} Jiaqi Chen, Bingqian Lin, Xinmin Liu, Lin Ma, Xiaodan Liang, and Kwan-Yee K. Wong. "Affordances-oriented planning using foundation models for continuous vision-language navigation." AAAI, 2025.

\bibitem{b30} Xuan Yao, Junyu Gao, and Changsheng Xu. "Navmorph: A self-evolving world model for vision-and-language navigation in continuous environments." ICCV, 2025.


\end{thebibliography}
\end{document}